\documentclass[a4paper,twoside]{article}
\usepackage{cite}
\usepackage{epsfig}
\usepackage{subcaption}
\usepackage{calc}
\usepackage{amsmath,amssymb,amsfonts}
\usepackage{amstext}
\usepackage{amsthm}
\usepackage{multicol}
\usepackage{pslatex}
\usepackage{apalike}
\usepackage{algorithm}
\usepackage{caption}
\usepackage{svg}
\usepackage{algpseudocode}
\usepackage{adjustbox}
\usepackage{tabularx}
\usepackage{booktabs}
\usepackage{graphicx}
\usepackage{textcomp}
\usepackage{xcolor}
\usepackage[column=O]{cellspace}
\usepackage[none]{hyphenat}
\usepackage{lipsum}
\usepackage{hyperref}
\usepackage[symbol]{footmisc}

\algnewcommand\algorithmicforeach{\textbf{for each}}
\algdef{S}[FOR]{ForEach}[1]{\algorithmicforeach\ #1\ \algorithmicdo}
\DeclareMathOperator*{\argmax}{arg\,max}

\usepackage{SCITEPRESS}

\begin{document}

\title{Real-time Active Vision for a Humanoid Soccer Robot Using\\
Deep Reinforcement Learning}

\author{\authorname{Soheil Khatibi\sup{1}\href{https://orcid.org/0000-0002-2968-3576}{\includegraphics[width=3mm]{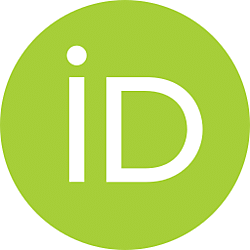}}, Meisam Teimouri\sup{1}\href{https://orcid.org/0000-0003-0465-6528}{\includegraphics[width=3mm]{orcid.jpg}}, Mahdi Rezaei\sup{2}\href{https://orcid.org/0000-0003-3892-421X}{\includegraphics[width=3mm]{orcid.jpg}}}
\affiliation{\sup{1}Mechatronics Research Laboratory, Qazvin Azad University, Qazvin, IR}
\affiliation{\sup{2}Institute for Transport Studies, University of Leeds, Leeds, UK}
\email{\{s.khatibi, m.teimouri\}@qiau.ac.ir, m.rezaei@leeds.ac.uk}
\vspace{-10mm}
}


\abstract{In this paper, we present an active vision method using a deep reinforcement learning approach for a humanoid soccer-playing robot.
The proposed method adaptively optimises the viewpoint of the robot to acquire the most useful landmarks for self-localisation while keeping the ball into its viewpoint.
Active vision is critical for humanoid decision-maker robots with a limited field of view. To deal with an active vision problem, several probabilistic entropy-based approaches have previously been proposed which are highly dependent on the accuracy of the self-localisation model.
However, in this research, we formulate the problem as an episodic reinforcement learning problem and employ a Deep Q-learning method to solve it.
The proposed network only requires the raw images of the camera to move the robot's head toward the best viewpoint.
The model shows a very competitive rate of $80\%$ success rate in achieving the best viewpoint. 
We implemented the proposed method on a humanoid robot simulated in Webots simulator.
Our evaluations and experimental results show that the proposed method outperforms the entropy-based methods in the RoboCup context, in cases with high self-localisation errors.
}

\onecolumn \maketitle \normalsize \setcounter{footnote}{0} \vfill

\vspace{-1mm}
\section{\uppercase{Introduction}}
\label{sec:introduction}
Active Vision, by definition, is the method of actively planning, manipulating, and adjusting the camera viewpoint, with the goal of obtaining the most optimised information from a given environment, which can be either static or dynamic.
Object/landmarks detection in an environment is an important research problem that has already been addressed and studied by many researchers such as in  \cite{rezaei2017computer}, \cite{teimouri2019real}.
However, in active vision, the goal is to reach the viewpoints which contain most important landmarks and objects so that there is a chance to detect them.
Speaking about the importance of active vision, there are lots of cases where we encounter with limited computational and processing resources. 
In such circumstances, the best plan is to try to use the existing resources in an optimised manner in order to make the most of it.
Another application of the active vision is when the field of view of the camera is limited.
In such cases, in order to gain the most possible observations, it would be wise to manipulate the viewpoint in a way that more observations are sensed.
Furthermore, in case of occlusions and partial visibility, using an active vision algorithm to intelligently conduct the eyesight toward more informative observations will be extremely helpful.

In order to plan for the best viewpoint of the camera, knowing the accurate state of the environment and the agent itself are crucial.
In active vision, due to dynamic nature of the objects and the uncertainty in the state of the agent, the state of the environment and the agent itself are not exactly available.
Therefore, an active vision system should model its state and the environment as accurately as possible.
This is one of the major challenges in this field.
On the other hand, trying to determine the best viewpoint with a high efficiency considering the limitations of an agent in a dynamic environment can be computationally expensive.
So active vision has been studied from different points of view and there are different approaches in this field.

One of the common techniques is the probabilistic based solutions.
In this kind of attitudes, the belief of the robot is modelled with a probability distribution.
Such techniques choose the action that minimises the uncertainty of the belief \cite{burgard1997active}.
In \cite{seekircher2010entropy} and \cite{czarnetzki2010real}, the belief models the position of the robot.
The concept of entropy is utilised as a measure indicating the rate of uncertainty in a probability distribution.
Therefore, the best action is the one that minimises the entropy of current belief.
This approach, however, has some weaknesses.
For example, in order to choose the next appropriate action, we have to calculate the model entropy by assuming all of the actions have already been completed.
In these circumstances, regardless of the fact that the state of the dynamic objects can not be estimated accurately, the performance of the model will be highly dependent on the accuracy of the current belief.
Another approach in active vision is to formulate the problem as a reinforcement learning where a big variety of modern algorithms can be utilised such as deep neural networks.
These approaches have witnessed great progress \cite{han2019active}, \cite{cheng2018reinforcement} in recent years thanks to advances in reinforcement learning and deep neural networks.
However, these methods are substantially in need of significant processing resources as well as time-consuming training procedures.
Also due to the complexity of the state and action space, training a robust model will be non-trivial and challenging.
As another challenge, and depending on the complexity of the task, usually performing a real-world training is not feasible and this should be accomplished in a simulator environment.
However, the trained models in the simulation environment can not necessarily perform in real-world environments with the same performance. 

As of the main contributions of this research, we first formulate the problem as an episodic reinforcement learning problem by defining a Markov Decision Process.
The objective is to adjust the head viewpoint through an optimum direction that will have the best observations so that the self-localisation reaches the highest accuracy, along with an accurate belief from the current state of the environment.
In each episode, the best action is determined using an entropy minimising methodology.
Since the training process is performed in the simulation environment, the position of the robot is accurately available.
So, the entropy minimisation method for determining the best action would be efficient.
The algorithm used for training is DDQN and PER \cite{mnih2015human}, \cite{van2016deep} and \cite{schaul2015prioritized}, which uses an experience replay memory to keep records of previous experiences so far.
As the second contribution in this research, we make the action selection process independent of the localisation error and the belief error of the environment.
The input of the algorithm is just the raw images and the output is a vector of q-values indicating how useful each action is.
Thus, having a trained model, we can control the head position by simply passing the current image to the model and select the appropriate action with the highest q-value without requiring any further input information.

We provide further details in the next sections.
The rest of the paper is organised as follows: In Section~\ref{sec:RelatedWork}, the main research-works and achievements accomplished in this field has been outlined.
In Section~\ref{sec:Methodology}, the problem is defined and our proposed method has been presented in details.
In Section~\ref{sec:experimental}, the outcome of the experimental results has been discussed.
And finally, the concluding remarks are provided in Section~\ref{sec:conclusion}.

\vspace{-3mm}
\section{\uppercase{Related Work}}
\label{sec:RelatedWork}
\vspace{-1mm}
Despite the fact that an active sensor transmits and receives information from the environment, active vision has been referred to as a planning strategy challenge to control the process of environment perception around an agent.
This can be accomplished by manipulating the agent or robot's viewpoint, although it does not transmit any information, but only receives some environmental information \cite{bajcsy1988active}.

It is now decades that this research field has been attracting interests for coping with challenges like occlusions, limited field of view, limited resolution of the camera, and low computational resources.
\cite{swain1993promising} outlines and explains some fields and research areas in active vision such as gaze control, hand-eye coordination, and embedding techniques in robotic settings.

More recently, \cite{chen2011active} summarised some of the developments of active vision in the robotic fields over the past 15 years, and describes problems arising from applications such as object recognition \cite{4648837}, tracking and search, localisation, and mapping.

\cite{mitchell2014active} has studied the visual behaviour of Marmosets such as their saccadic behaviours in different situations considering their neurophysiology and neural mechanisms.
From the cognitive point of view, \cite{kieras2014towards} surveys some advances in cognitive models for human-computer interaction using active vision systems.
Also \cite{rolfs2015attention} investigates the perceptual and cognitive processes and their relations with the memory from the psychological point of view.

Very recently, \cite{ammirato2017dataset} has provided an active vision dataset including bounding boxes of some detected objects which can be used not only in object detection tasks, but also in active vision methods as a simulated environment.

\cite{falanga2017aggressive} applies active vision for manipulating quadrotor robot orientation in the task of passing narrow gaps.

In RoboCup domain, Seekircher et al. \cite{seekircher2010entropy} have suggested a method for actively sensing the environment in order to increase the efficiency of self-localisation and ball tracking by minimising the entropy of an underlying particle distribution.
They have demonstrated the approach on a humanoid robot on a RoboCup soccer field. 

Czarnetzki et al \cite{czarnetzki2010real} have done a similar work in this field which provides optimal decisions based on the current localisation and its uncertainty in a soccer robot scenario.

Mattamala et al. \cite{mattamala2015dynamic} propose a method for mapping both static and dynamic information to an action as the best head manoeuvre action.
The method determines the appropriated head action using  on a combined score  based on the existing obstacles in the scene, the cost of performing each action, and the limitation ahead.
The method performs real-time.

Han et al. \cite{han2019active} formulates active object detection as a sequential action decision process and propose a duelling architecture to resolve it.
It uses an object detector to detect and localise the object in the image and feeds the image along with a representation of the detected object to a deep q-network.
The model outputs the state-action values in order to reach a higher rate of object detection.

Cheng et al. \cite{cheng2018reinforcement} also use an object detector module for both actor and critic networks in their actor-critic architectures in situations where the camera is active.
They propose hand/eye controllers that learn to move the camera to keep the object within the visible field of view, in coordination to manipulating it to achieve the desired goal in cluttered or occluded environments.
Both recent works require an object detector.
Although this may help and accelerate the course of learning, it can be unavailable in some circumstances and environments.

\vspace{-3mm}
\section{\uppercase{Methodology}}
\label{sec:Methodology}
\vspace{-1mm}
We are interested in reaching the best viewpoint to perceive the most useful observations from the environment for self-localisation while having an object of interest within the viewpoint of the Robot.
First let's provide more details on the task specifications. 

\vspace{-1mm}
\subsection{Task Specifications}
\label{sec:TS}
\vspace{-1mm}
In a humanoid robot, the viewpoint is controlled by the actuators of the robot's head and the environment is a Robocup humanoid soccer field.
The important object of interest in the humanoid soccer field is the ball and the observations are in fact the landmarks that contribute to the process of the robot localisation.
Therefore the goal is not only to use perceived observations to localise the robot in a dynamic environment, but also to be in control of the match as the consequence of always having the ball in the viewpoint.

We see this as a sequential decision-making problem; therefore, to cope with the challenge, we offer an episodic reinforcement learning solution, with two main actors: the agent and a dynamic environment.
Environment refers to the soccer field with its whole components that the agent should interact with, and the agent is the humanoid robot which aims to learn the optimal policy to optimise the robot's viewpoint.
The agent-environment interaction is formulated as a Partially Observable Markov Decision Process (POMDP) represented by observations $o \in \mathcal{O}$, states $s \in \mathcal{S}$, actions $a \in \mathcal{A}$, and the reward function $r:S \times A \rightarrow \mathbb{R}$.
The components of this POMDP are described below:

\begin{itemize}
\item Observations: Visual Information has diversely been used in sequential decision-making problems so far in different forms.
Here, observation $o_t$ is a gray-scale image captured at time step $t$ by a camera mounted on the head of the robot.
\item State representation: In this problem, states are represented as a sequence of observations $o_t$ that the robot sees.
The size of the stack can vary according to the circumstances of the problem.
\item Action Space: The action space is a set of discrete actions in which each action $a_t$ moves the robot viewpoint in a specific direction to a little fixed extent.
\item Reward function: The goal of our reward function is to encourage the robot to move its head through the goal position and penalises the robot from moving its head away from the goal position, alongside considering a negative reward for missing the ball.
This function is specified as below:
\begin{equation}
  Reward = 
  \begin{cases}
    -2, & \text{for missing all balls}\\
    \text{sign} (D'-D), & \text{elsewhere}
  \end{cases}
\end{equation}
where $D$ is the distance of head to the goal position before taking the action and $D'$ is the same distance after taking the action.
\end{itemize}

\vspace{-2mm}
\subsection{Goal Determination}
\label{sec:GD}
\vspace{-2mm}
At the beginning of each episode, the best possible viewpoint should be determined from the position of the robot and the ball.
As specified in the previous sub-section, the goal position is the viewpoint containing the landmarks that improve the self-localisation model of the robot in the best way while keeping the ball in it.
To accomplish this objective, we discretise the camera position space and evaluate all possible positions to determine the goal position.
Every camera position $p$ is represented by the tuple
\begin{equation}
p = (\theta_{pan}, \theta_{tilt})   
\end{equation}
\noindent where 
\begin{equation*}
\frac{-\pi}{2} < \theta_{pan} < \frac{\pi}{2}
\end{equation*}
\noindent and 
\begin{equation*}
\frac{\pi}{36} < \theta_{tilt} < \frac{13 \pi}{36} 
\end{equation*}
\noindent are the pan and title angles of the head actuators, which are discretised into 10 and 4 points, respectively.
Note that every camera position corresponds to a viewpoint, so we can use these terms interchangeably.

We represent the belief of the robot position using a multivariate normal distribution:

\begin{equation}
X \sim \mathcal{N}(\mu,\,\Sigma)    
\end{equation}

\noindent where $\mu$ is the mean vector representing the position of the robot and $\Sigma$ is the covariance matrix that shows the uncertainty related to the position.

To evaluate the efficiency of a viewpoint for self-localisation we employed an entropy-based method.
The best viewpoint is one that contains the ball and minimises the entropy of the Gaussian model.
The process of finding the best viewpoint is shown in Algorithm \ref{alg:ViepointExploration}.
The algorithm takes the robot and ball positions as input and returns the best viewpoint.

As shown in lines 2-11 of the algorithm, for each viewpoint $p$, we compute the expected entropy and update the best viewpoint $p^*$.
First, all observations measured from visible landmarks at the current position of the robot and viewpoint $p$ are determined in line 4. Each observation $z \in Z$ is related to a visible landmark and represents the distance and angle of the landmark to the robot.

\begin{figure}[t!]
\centering
\includegraphics[width=0.4\textwidth]{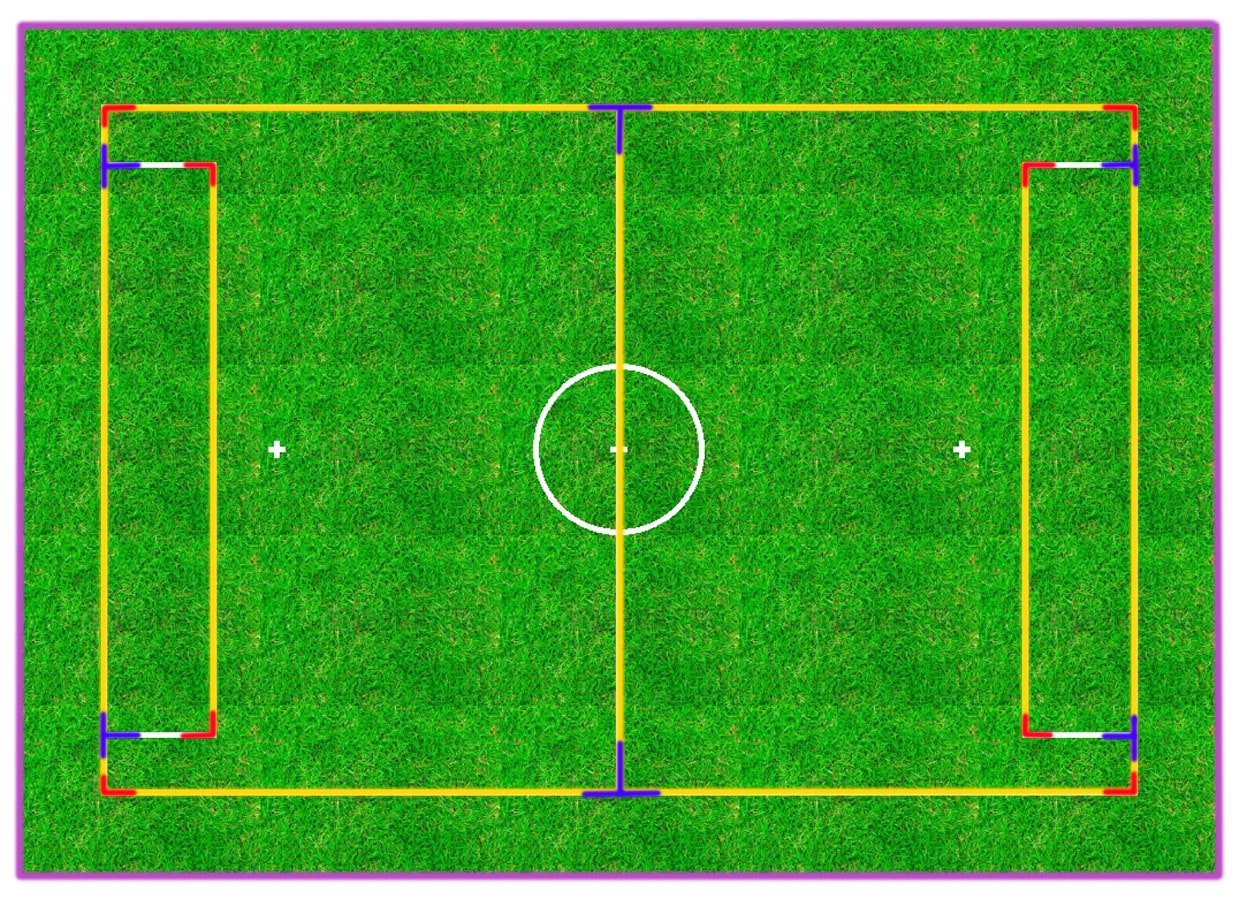}
\caption{Different parts of the soccer field markings, categorised and colour coded based on their types.
Red for "L"s, blue for "T"s, yellow for lines, and purple for the field boundary.}
\label{fig:Field}
\end{figure}

The visible landmarks can be determined easily from the field model.
Figure~\ref{fig:Field} illustrates different landmarks in the field categorised by different colours.
However, the measurements of landmarks are assumed to be noisy.
So each landmark that is projected to the viewpoint $p$ and located at a predetermined distance to the robot is considered as visible.
This distance is estimated experimentally for each type of landmark.
In \cite{seekircher2010entropy}, a visibility model is learned using a neural network.
Then for every $z$ we update the belief $X$ using Unscented Kalman filter (line 6) \cite{teimouri2016hybrid}).
The entropy for updated belief $X'$ is calculated in line 8.

\begin{algorithm}
\caption{viewpoint exploration}\label{alg:ViepointExploration}
\hspace*{\algorithmicindent} \textbf{Input:} $ X_t = \mathcal{N}(\mu_t,\,\Sigma_t)$ , $ballpose = (x_t, y_t)$
\begin{algorithmic}[1]
\State $H_{min} = \infty$
\ForEach {$p \in P$}
\State $X' = X_t$
\State $Z$ = $get\_observations(X', p)$
\ForEach {$z \in Z$}
\State $X' = apply\_UKF(X', z)$
\EndFor
\State $H_{X'}$ = $\frac{1}{2} \ln{(|(2\pi e)\Sigma|)}$
\If{ $H_{X'} < H_{min}$ and \\ \hspace{5mm} $ball\_is\_visible(X', ballpose)$}
\State $H_{min} = H_{X'}$
\State $p^{*} = p$
\EndIf
\EndFor
\State \Return $p^*$
\end{algorithmic}
\end{algorithm}

\begin{figure*}[ht!]
\vspace{-4mm}
\centering
\includegraphics[width=1\textwidth]{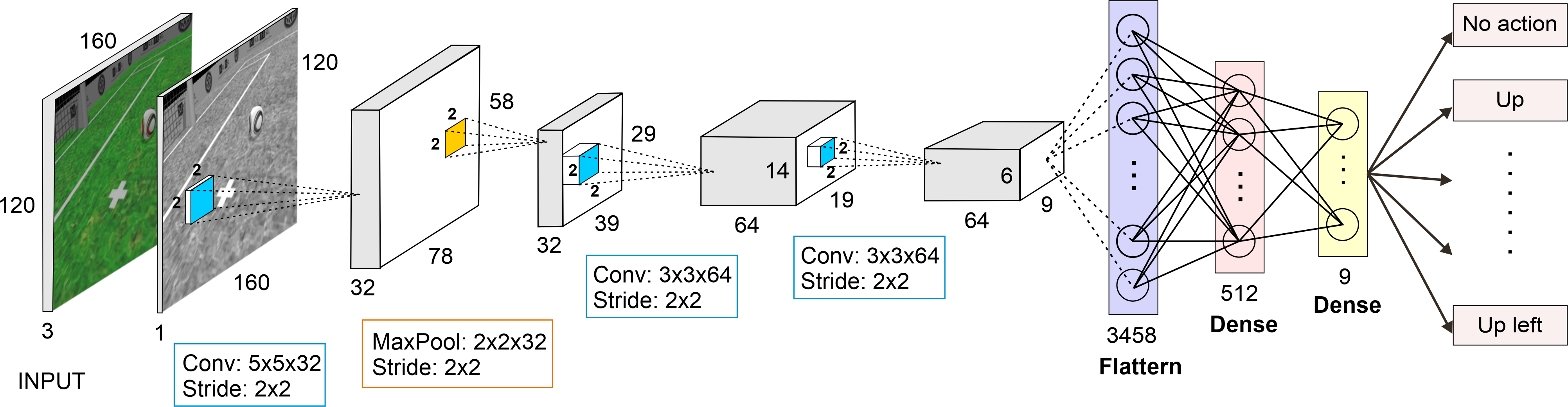}
\vspace{-3mm}
\caption{Neural network architecture. Note that ReLU activation function has been used for all layers except for the max-pooling layer and the last layer.}
\vspace{-1mm}
\label{fig:CNN}
\end{figure*}

\vspace{-3mm}
\subsection{Training Process and Details}
\label{sec:TPD}
\vspace{-1mm}
\cite{seekircher2010entropy} has used the output of an entropy-based algorithm in online controlling of the robot's head. This can lead to wrong results in the case of inaccurate localisation which is a common issue in localisation algorithms (such as \cite{seekircher2010entropy} and \cite{teimouri2016hybrid}). 

In our proposed sequential decision-making method, the input is the current image which is independent to the localisation accuracy.
However, this makes the task more complex and tricky in some specific positions due to the symmetry of the soccer field.

As mentioned earlier, this is an episodic task where at the beginning of each episode, the ball and robot are randomly positioned in the field with a random camera position.
Then the goal (i.e. the best viewpoint) is determined using the proposed algorithm in the previous section. 
Afterwards, the agent starts to take actions until termination. Note that the episode finishes successfully if the agent reaches the goal viewpoint and will fail if the agent misses the ball from its field of view. The episode also terminates after 20 time steps to avoid long episodes. After termination, the environment resets and another episode begins.

The Algorithm that has been applied to solve the represented problem is DDQN Algorithm \cite{van2016deep}, which is an extension of DQN \cite{mnih2015human}. Also Prioritised Experience Replay \cite{schaul2015prioritized} has been utilised. In DQN algorithms, $Q$-function is approximated using a deep neural network. $Q$-values are real numbers indicating the quality of each action in a specific state. More formally, $Q(s,a)$ is the expected cumulative discounted reward after taking action $a$ in state $s$.
In double deep $Q$-learning, selection and evaluation are untangled. One network parametrised with $\theta$ is used for selecting the action and another one parametrised with $\theta'$ is used to evaluate $Q$-values.
The $Q$-values should be updated towards a target value:

\begin{equation}
{Y_t}^{DoubleQ} \equiv R_{t+1}+{\gamma}Q(S_{t+1},\argmax_{a} Q(S_{t+1},a;\theta_t);\theta'_t)
\end{equation}
Each experience $e_t = (s_t, a_t, r_t, s_{t+1})$ consists of state, action, the reward after taking the action, and the subsequent state after taking the action.
In the training phase, different batches are picked to train the neural network from a prioritised experience replay in which important transitions are picked more frequently.

The architecture of the convolutional neural network is shown in Figure~\ref{fig:CNN}.
It takes gray-scale images as input and outputs a vector of $Q$-values whose length equals the number of possible actions.
We fed a tensor with the size of $160\times120\times 1$ as the input to the network.
The first hidden layer convolves 32 filters of $5\times5$ with Stride 2 and then applies a non-linear activation function.
The second hidden layer applies a Max pooling of $2\times2$ with stride 2.
The third hidden layer convolves 64 filters of $3\times3$ with stride 2, followed by a non-linear activation function.
This is then followed by another convolutional layer that convolves 64 filters of $3\times3$ with stride 2 followed by a non-linear activation function.
Then there is a hidden fully-connected layer that consists of 512 units.
Finally, the output layer is a fully-connected linear layer with an output for each action.

The hyperparameters used in our work are shown in table~\ref{tab:table1}.
The Adam optimiser is used for the training phase of the network.
Also the learning rate is constant during the whole training steps.
\begin{table}[ht]
\caption{HyperParameters}
\label{tab:table1}
\centering
\begin{tabular}{|l|l|}
  \hline
  \textbf{Hyperparameter} & \textbf{Value}\\
  \hline
  Minibatch size & $32$ \\
  Replay memory size & $1000000$\\
  Learning rate & $0.0005$\\
  Discount factor & $0.99$\\
  Target network update frequency & $10000$\\
  Initial exploration rate & $1$\\
  Final exploration rate & $0.02$\\
  Prioritised replay buffer $\alpha$ & $0.6$\\
  Initial prioritised replay buffer $\beta$ & $0.4$\\
  \hline
\end{tabular}
\end{table}
\begin{figure*}[ht!]
\vspace{-4mm}
\centering
\begin{subfigure}{.4\linewidth}
\centering
\includegraphics[width=.7\linewidth]{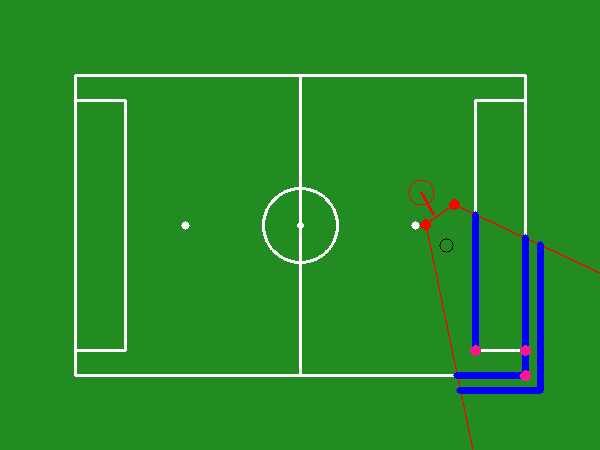}
\label{sfig:testa}
\end{subfigure}%
\begin{subfigure}{.4\linewidth}
\centering
\includegraphics[width=.7\linewidth]{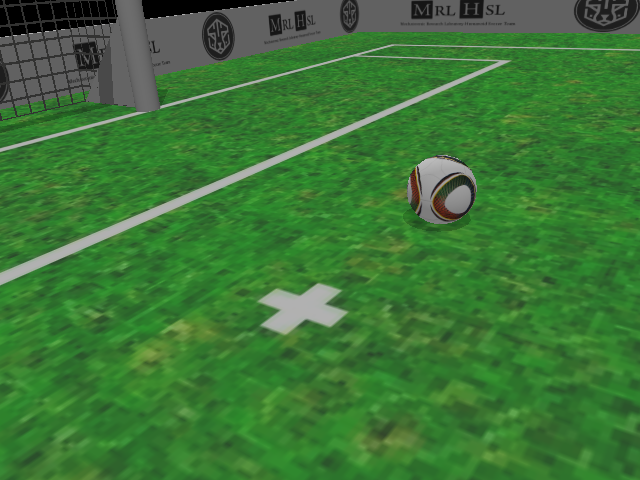}
\label{sfig:testb}
\end{subfigure}\par\medskip
\begin{subfigure}{.2\linewidth}
\centering
\includegraphics[width=.9\linewidth]{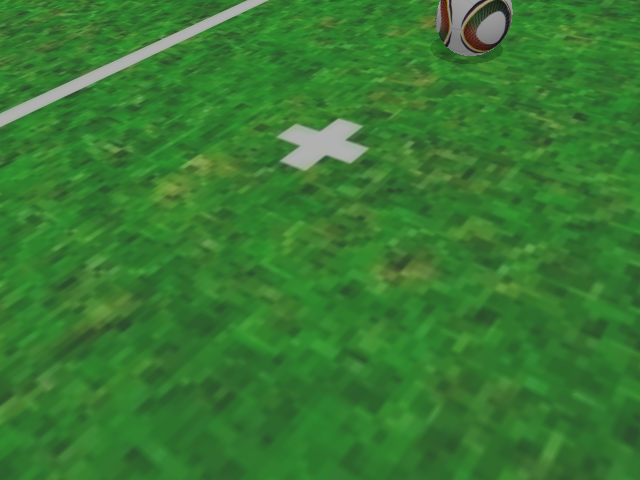}
\label{sfig:testc}
\end{subfigure}%
\begin{subfigure}{.2\linewidth}
\centering
\includegraphics[width=.9\linewidth]{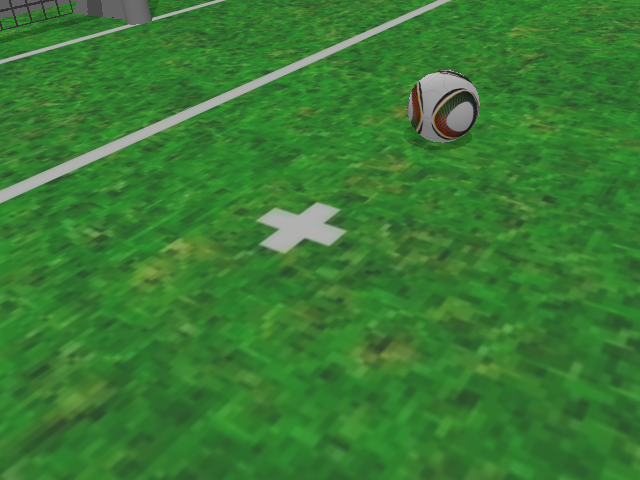}
\label{sfig:testd}
\end{subfigure}%
\begin{subfigure}{.2\linewidth}
\centering
\includegraphics[width=.9\linewidth]{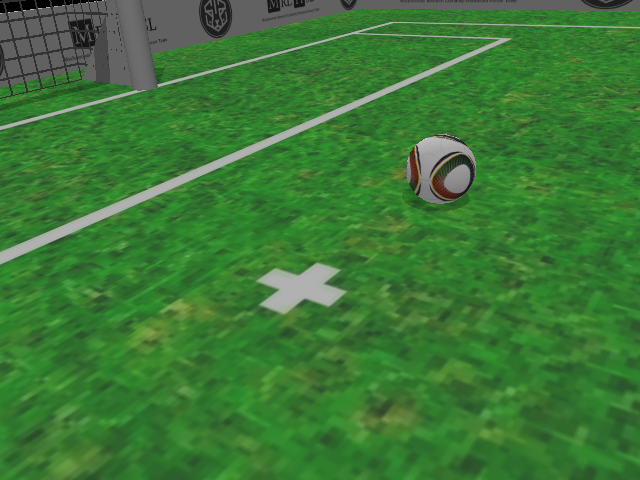}
\label{sfig:teste}
\end{subfigure}%
\begin{subfigure}{.2\linewidth}
\centering
\includegraphics[width=.9\linewidth]{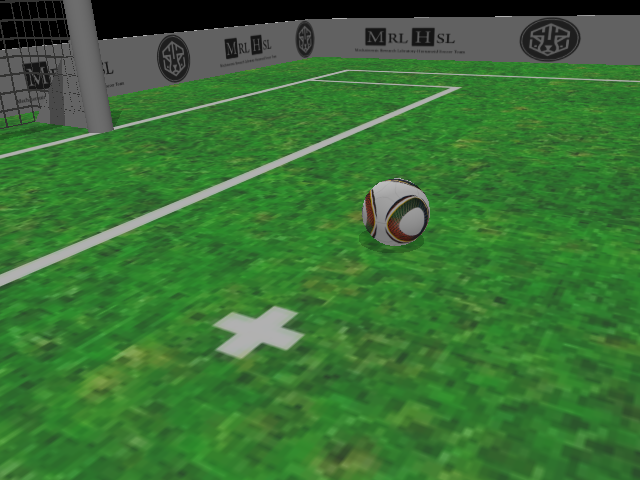}
\label{sfig:testf}
\end{subfigure}\par\medskip
\begin{subfigure}{.2\linewidth}
\centering
\includegraphics[width=.9\linewidth]{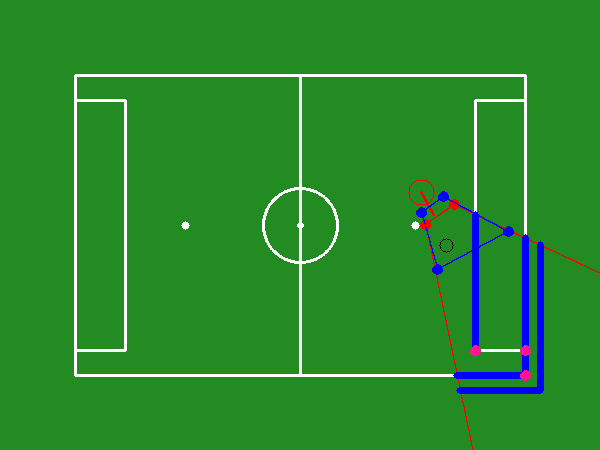}
\label{sfig:testg}
\end{subfigure}%
\begin{subfigure}{.2\linewidth}
\centering
\includegraphics[width=.9\linewidth]{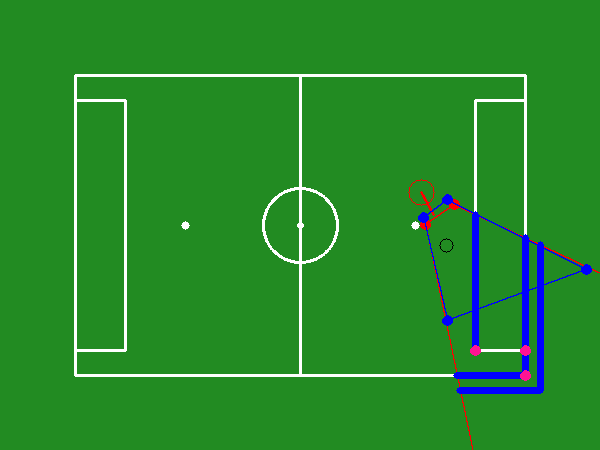}
\label{sfig:testh}
\end{subfigure}%
\begin{subfigure}{.2\linewidth}
\centering
\includegraphics[width=.9\linewidth]{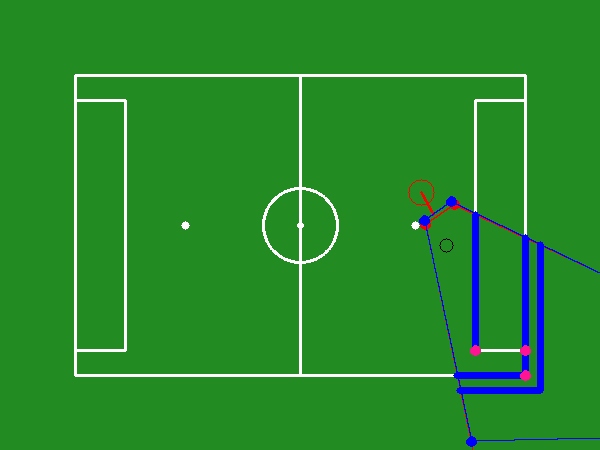}
\label{sfig:testi}
\end{subfigure}%
\begin{subfigure}{.2\linewidth}
\centering
\includegraphics[width=.9\linewidth]{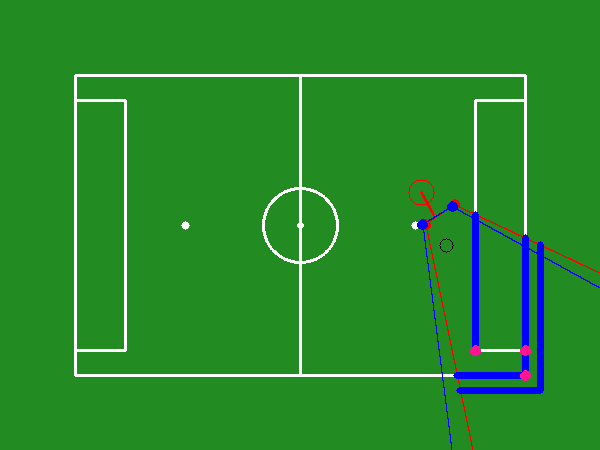}
\label{sfig:testj}
\end{subfigure}\par\medskip
\begin{subfigure}{.2\linewidth}
\centering
\includegraphics[width=.9\linewidth]{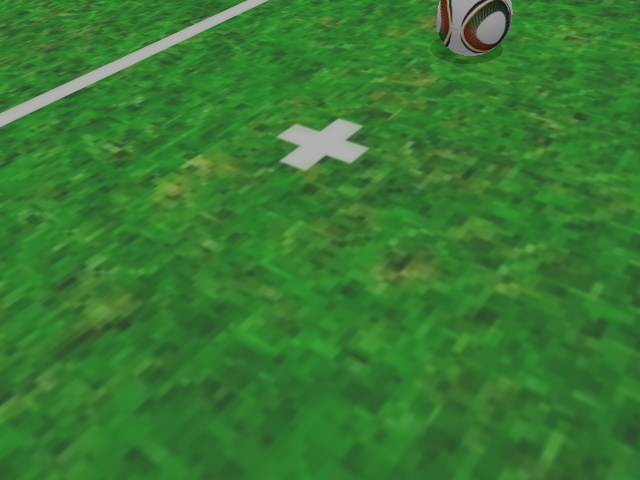}
\label{sfig:testk}
\end{subfigure}%
\begin{subfigure}{.2\linewidth}
\centering
\includegraphics[width=.9\linewidth]{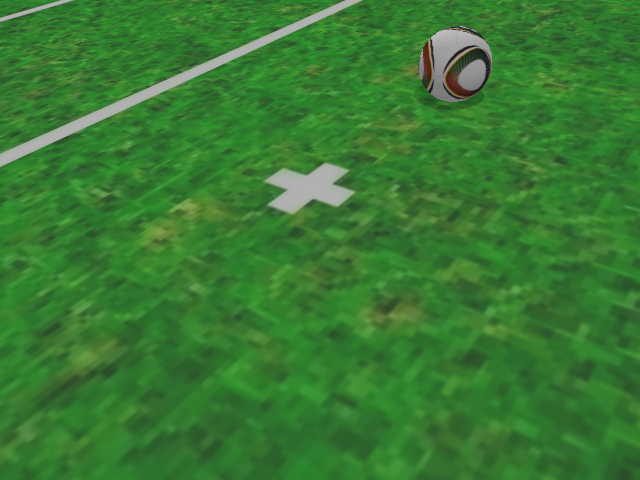}
\label{sfig:testl}
\end{subfigure}%
\begin{subfigure}{.2\linewidth}
\centering
\includegraphics[width=.9\linewidth]{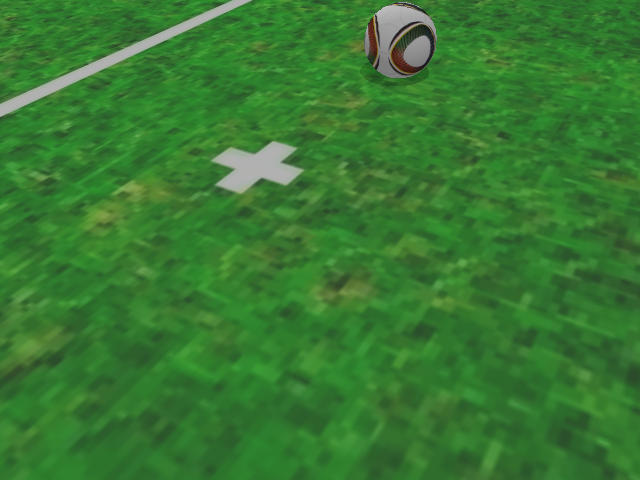}
\label{sfig:testm}
\end{subfigure}%
\begin{subfigure}{.2\linewidth}
\centering
\includegraphics[width=.9\linewidth]{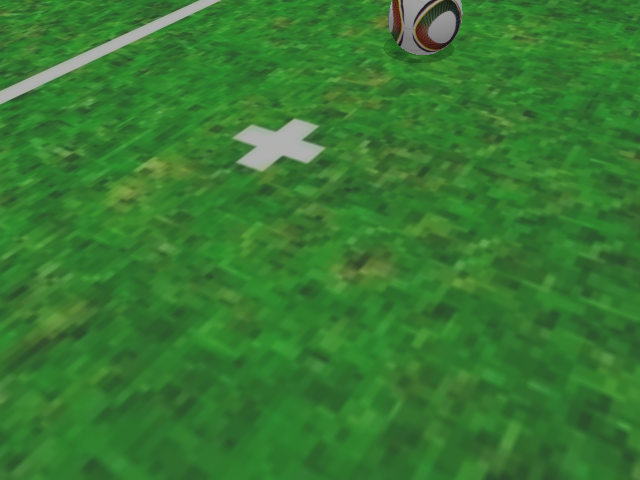}
\label{sfig:testn}
\end{subfigure}\par\medskip
\begin{subfigure}{.2\linewidth}
\centering
\includegraphics[width=.9\linewidth]{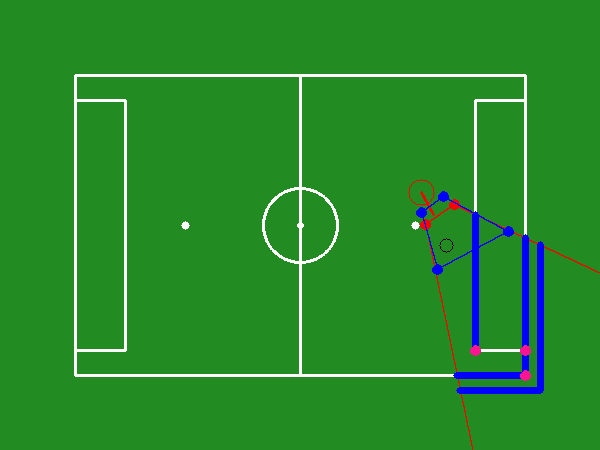}
\label{sfig:testo}
\end{subfigure}%
\begin{subfigure}{.2\linewidth}
\centering
\includegraphics[width=.9\linewidth]{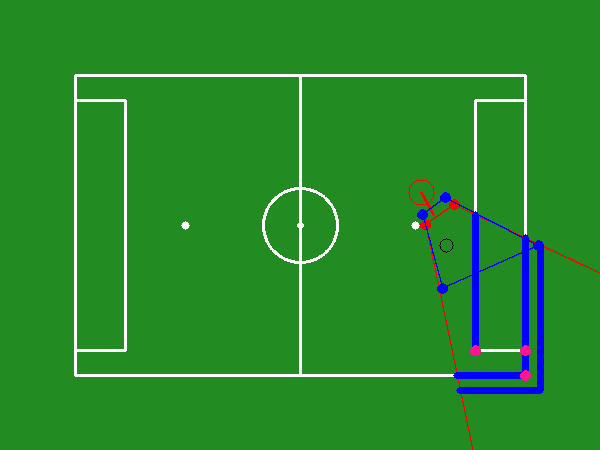}
\label{sfig:testp}
\end{subfigure}%
\begin{subfigure}{.2\linewidth}
\centering
\includegraphics[width=.9\linewidth]{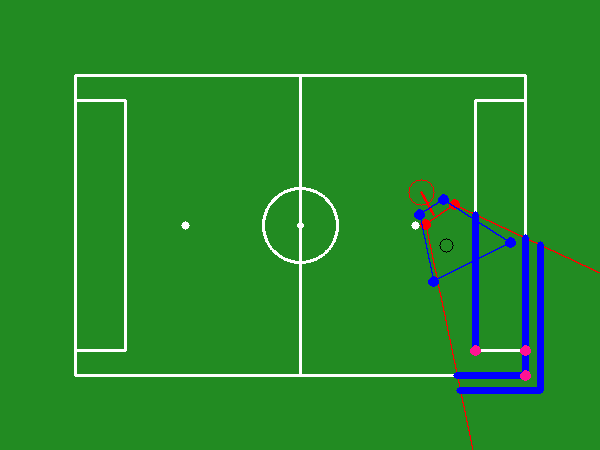}
\label{sfig:testq}
\end{subfigure}%
\begin{subfigure}{.2\linewidth}
\centering
\includegraphics[width=.9\linewidth]{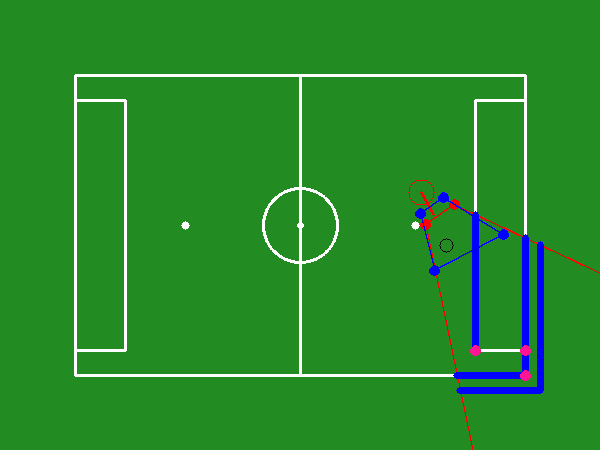}
\label{sfig:testr}
\end{subfigure}
\vspace{1mm}
\caption{
From top to bottom, Row 1: The best viewpoint of the robot (right) from specified position (left).
Row 2, 3: Active view point adjustment and visible landmarks started from left.
Row 4, 5: Random viewpoint adjustment and visible landmarks.
Note that the robot's position is marked with a red circle. the red polygons show the best viewpoint and the blue ones show the current viewpoint. Also visible landmarks in the best viewpoint are illustrated with blue lines and pink points.}
\label{fig:test}
\vspace{-2mm}
\end{figure*}
\vspace{-4mm}
\section{Experiments and Results}
\label{sec:experimental}
\vspace{-1mm}
The video link \url{https://youtu.be/kOX_vY6ir5M} represents the summary of our experiments.

This section is organised as follows. In Sec.~\ref{sec:ExperimentsEnvironment}, we introduce the Reinforcement Learning (RL) environment we used for the experiments and our training procedure, as well as the specs of the agent that has been trained.
In Sec.~\ref{sec:ShowCase}, we assess the performance of the proposed method during the training phase.
In Sec.~\ref{sec:Comparison}, we check how the self-localisation error affects the performance of the proposed method against the entropy-based method.

We evaluate the proposed DQN through some experiments and compare the results with previous works in RoboCup competitions.
While many types of experiments can be considered to assess, in this research, we limited our experiments on some of the most important aspects such as the success rate in reaching the best viewpoint, number of steps taken to reach the best viewpoint, number of steps taken to miss the ball from the viewpoint, and comparing the proposed method with the entropy-based method when the self-localisation error increases.
Figure~\ref{fig:test} shows the operation of the proposed method in an example robot and camera position, compared to random action selection.

\subsection{Experiments Environment}
\label{sec:ExperimentsEnvironment}
\vspace{-2mm}
Our training environment is designed and set up using the Webots simulator \cite{michel2004cyberbotics}.
We used Tensorflow \cite{abadi2016tensorflow} and Stable-baselines \cite{stable-baselines}, an improved version of OpenAI Baselines \cite{baselines}, to implement DDQN and PER.
Position commands are used to control the head joints at each time step.
Each command takes 320ms to rotate the camera viewpoint by approximately 3 degrees.
All processes such as simulating, training, and inference are run in a PC platform equipped with an Intel Core i7- 7700HQ processor, and an Nvidia GeForce GTX 1060 GPU.
Note that the average time of inference on this PC was 0.0023 seconds which can be used in real-time.

\begin{figure}[t!]
\centering
\includegraphics[width=0.45\textwidth]{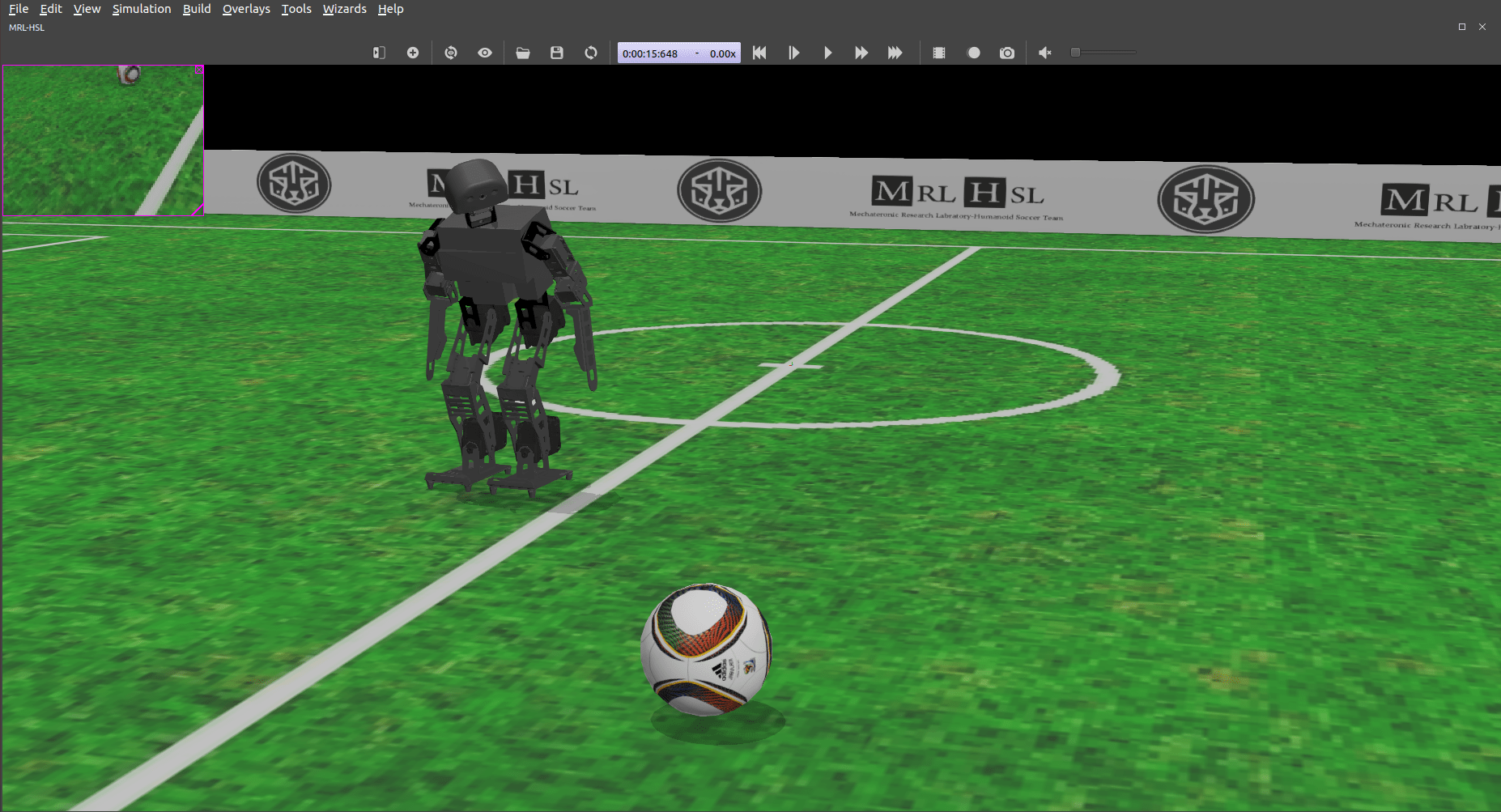}
\caption{An overview of the simulator and training environment.}
\label{fig:Env}
\end{figure}

We tested the proposed method on the MRL-HSL humanoid robot \cite{mahmoudi2019mrl} simulated in Webots.
An overview of the simulated environment is shown in Figure~\ref{fig:Env}.
The goal of the agent is clearly to reach the camera position in which the best landmarks are visible from any initial random position by manipulating neck and head joints.
All episodes are bound to 20 time steps but may terminate prematurely in the case of missing the ball from the field of view (as failure) or reaching the best viewpoint (as success).

The agent controls the camera viewpoint within a certain limited region.
The camera actions, include rotations to a specific extent (3 degrees in our case) and in a certain directions, as described in section \ref{sec:TS}.
The other joint angles of the robot remain same as the starting situation.
When the camera position is within a specific range of tolerance from the goal position, the episode ends with success.

\vspace{-1mm}
\subsection{Model Performance}
\label{sec:ShowCase}
\vspace{-1mm}

\begin{figure}[t!]
\centering
\includegraphics[width=0.45\textwidth]{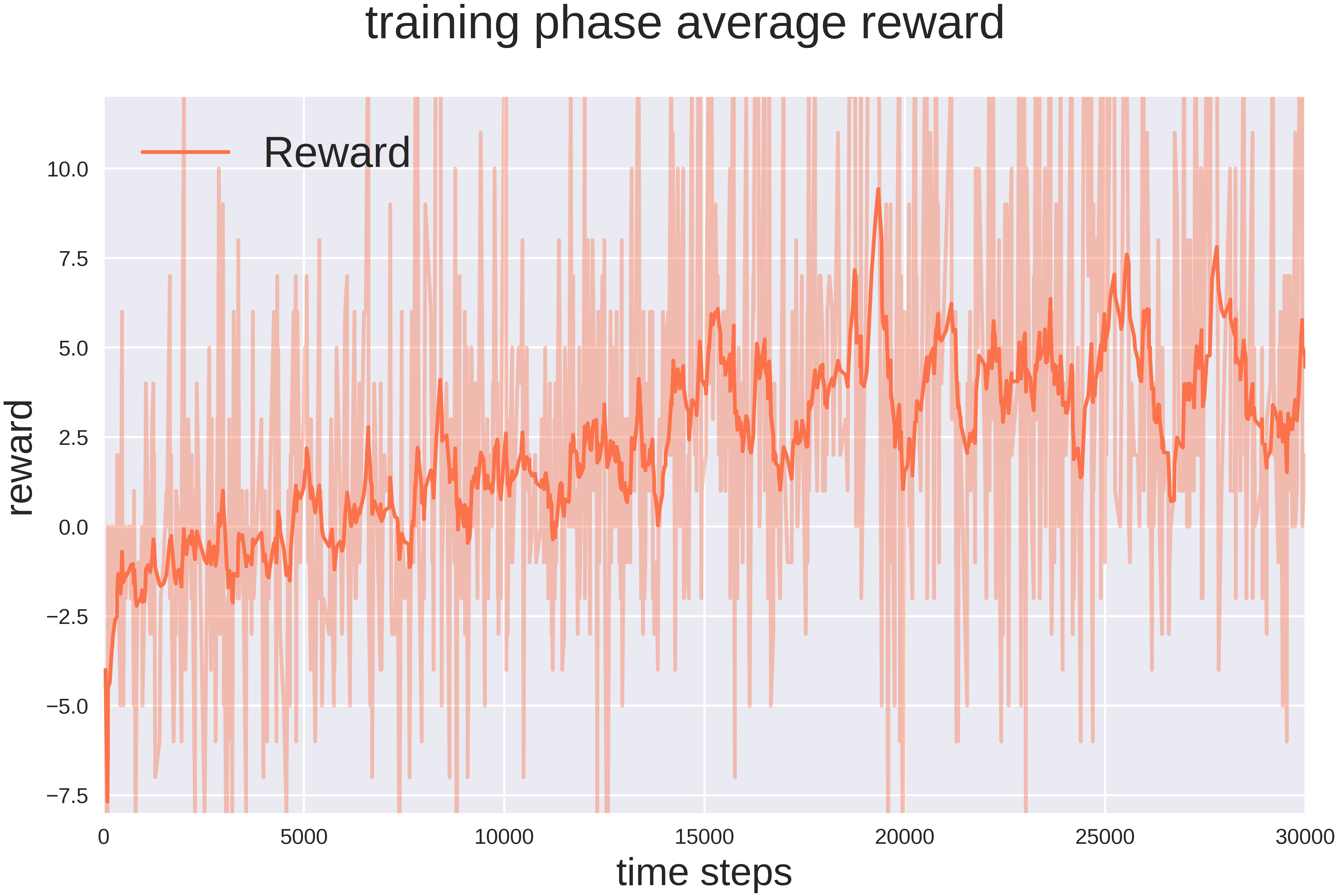}
\caption{Total reward gained per episode during the training course.}
\label{fig:epRe}
\end{figure}

\begin{figure}[t!]
\centering
\includegraphics[width=0.45\textwidth]{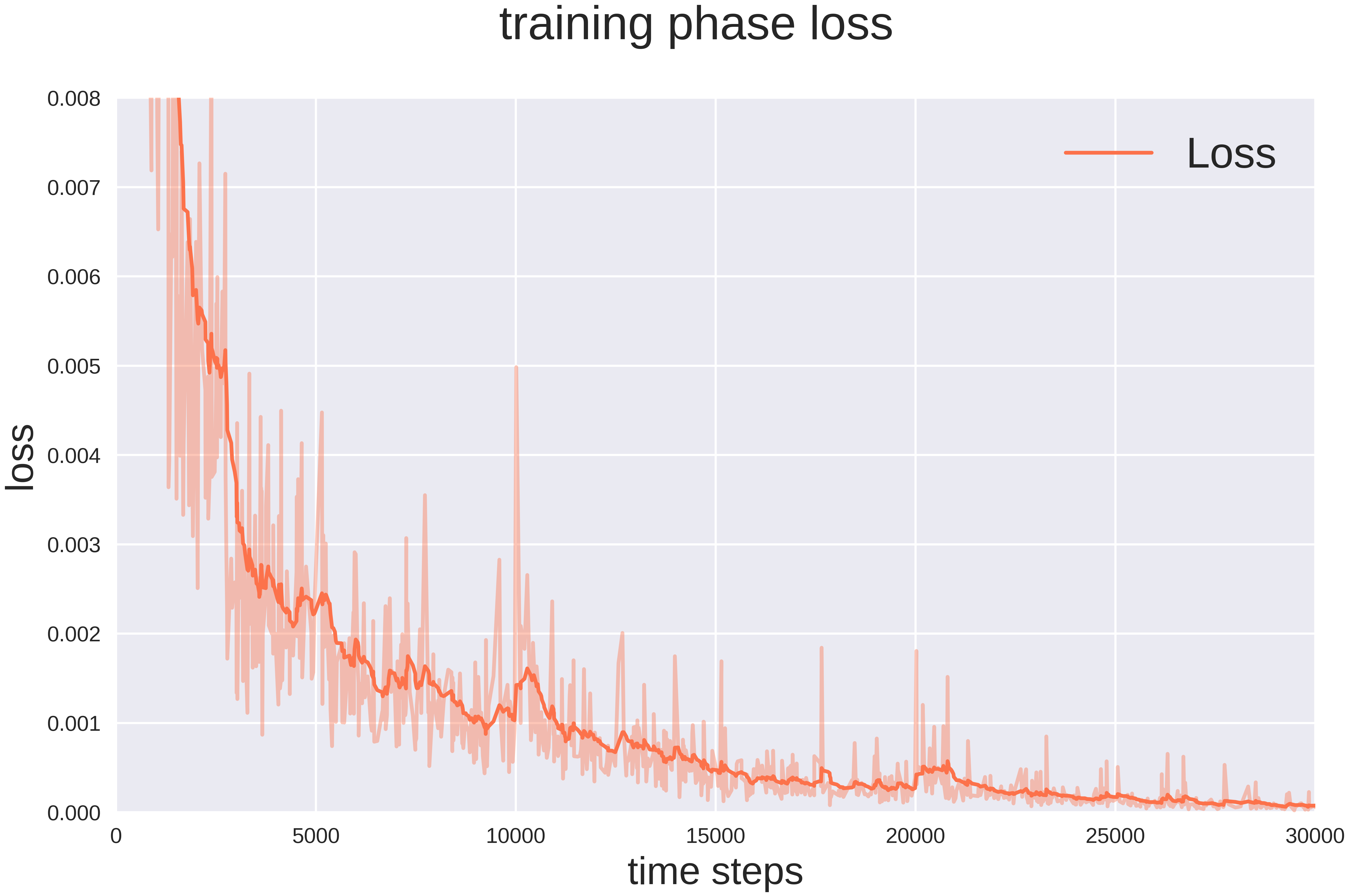}
\caption{Loss function of the model during the training course.}
\label{fig:L}
\end{figure}

The agent has been trained for 30000 time steps.
Total reward per episode and loss function during training phase are illustrated in Figure~\ref{fig:epRe} and \ref{fig:L} respectively.
To evaluate the performance of the model, we propose and measure 3 criteria during the training phase.
The first criterion is Success rate.
In an episode, we define Success rate as
\begin{equation}
SuccessRate = \frac{|\{observed\;landmarks\}|}{|\{desired\;landmarks\}|}
\end{equation}
which indicates how much of the desirable landmarks in the best viewpoint are observed.
Since we know the accurate position of the robot in simulation environment, we can determine the best viewpoint using algorithm~\ref{alg:ViepointExploration}.
So the Success rate can be calculated by dividing the current number of observed landmarks to the number of visible landmarks in the best viewpoint.
The second criterion is Success duration which is the time steps taken to reach the goal.
If the robot doesn't reach the goal in that episode, it will be 20 indicating not to reach the goal.
Finally, the third criterion(which was helpful for us in the Ball loss rate reduction), is Ball loss duration which is the time steps taken to lose the ball in each episode.
If the robot doesn't lose the ball, it will be 20, indicating that the robot has not lost the ball in that episode.

\begin{figure}[t!]
\centering
\includegraphics[width=0.45\textwidth]{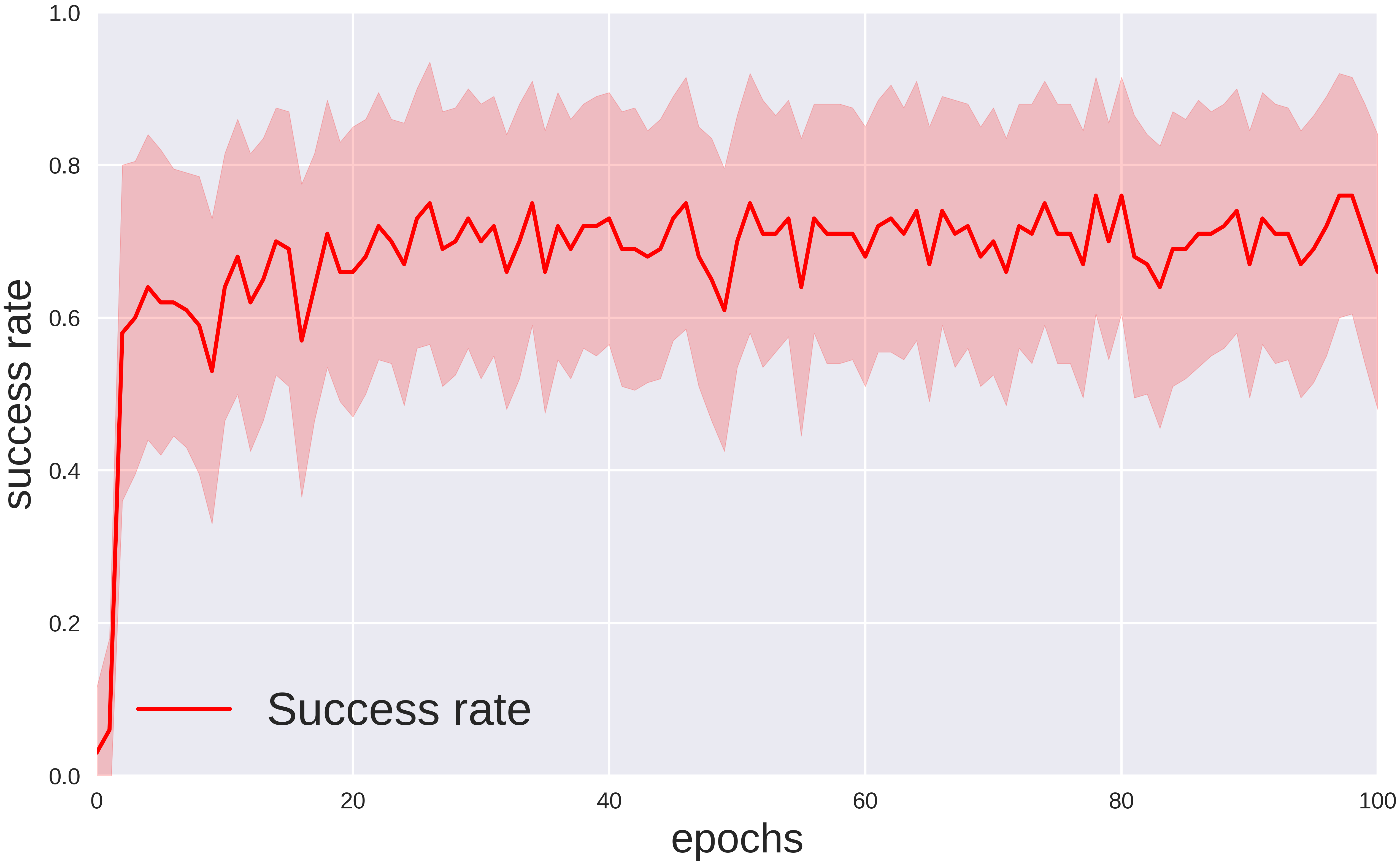}
\caption{\label{fig:successrate}The average Success rate measured during training epochs.}
\end{figure}

\begin{figure}[t!]
\centering
\includegraphics[width=0.45\textwidth]{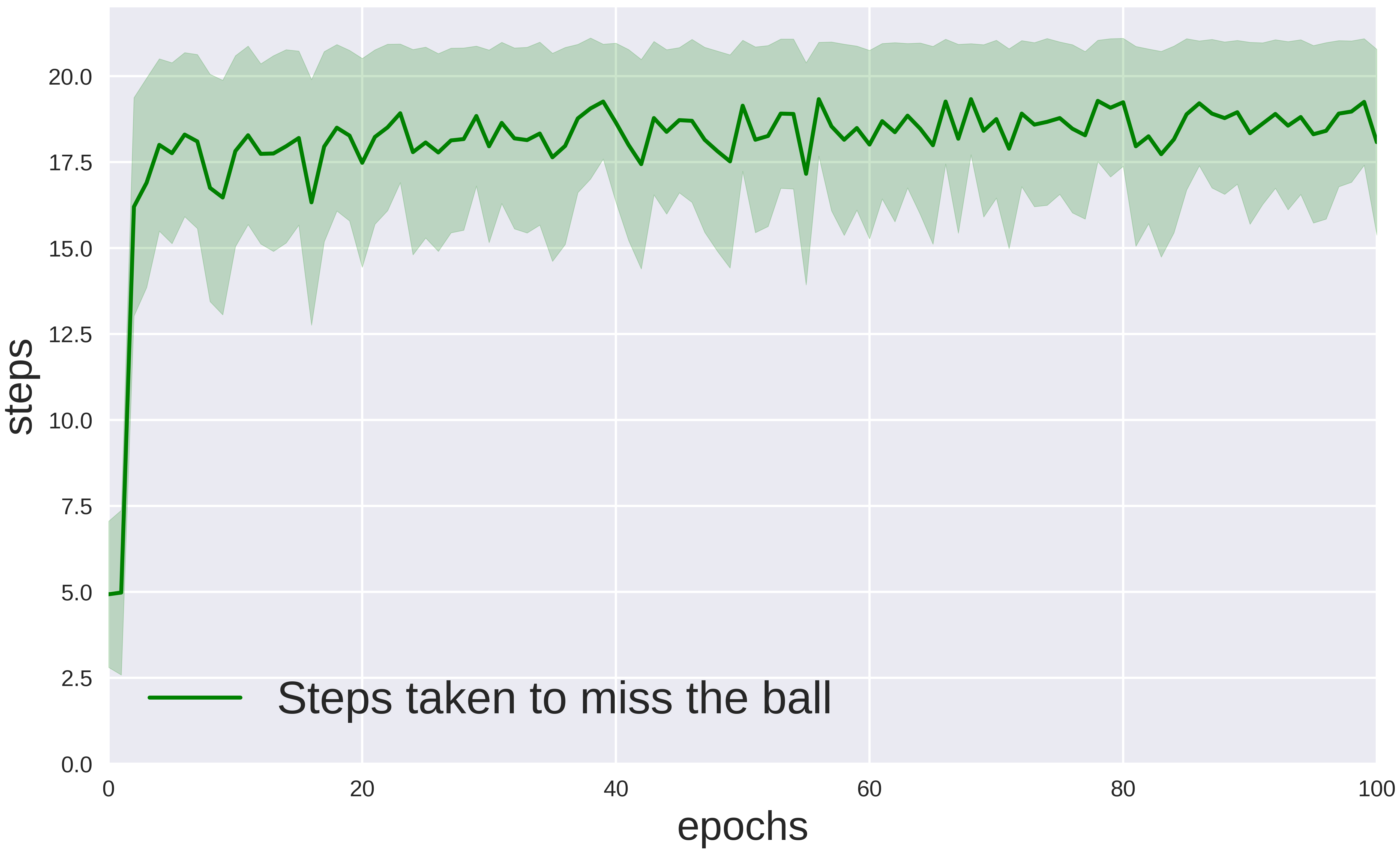}
\caption{\label{fig:miss}The average Ball loss duration measured during training epochs.}
\end{figure}

\begin{figure}[t!]
\centering
\includegraphics[width=0.45\textwidth]{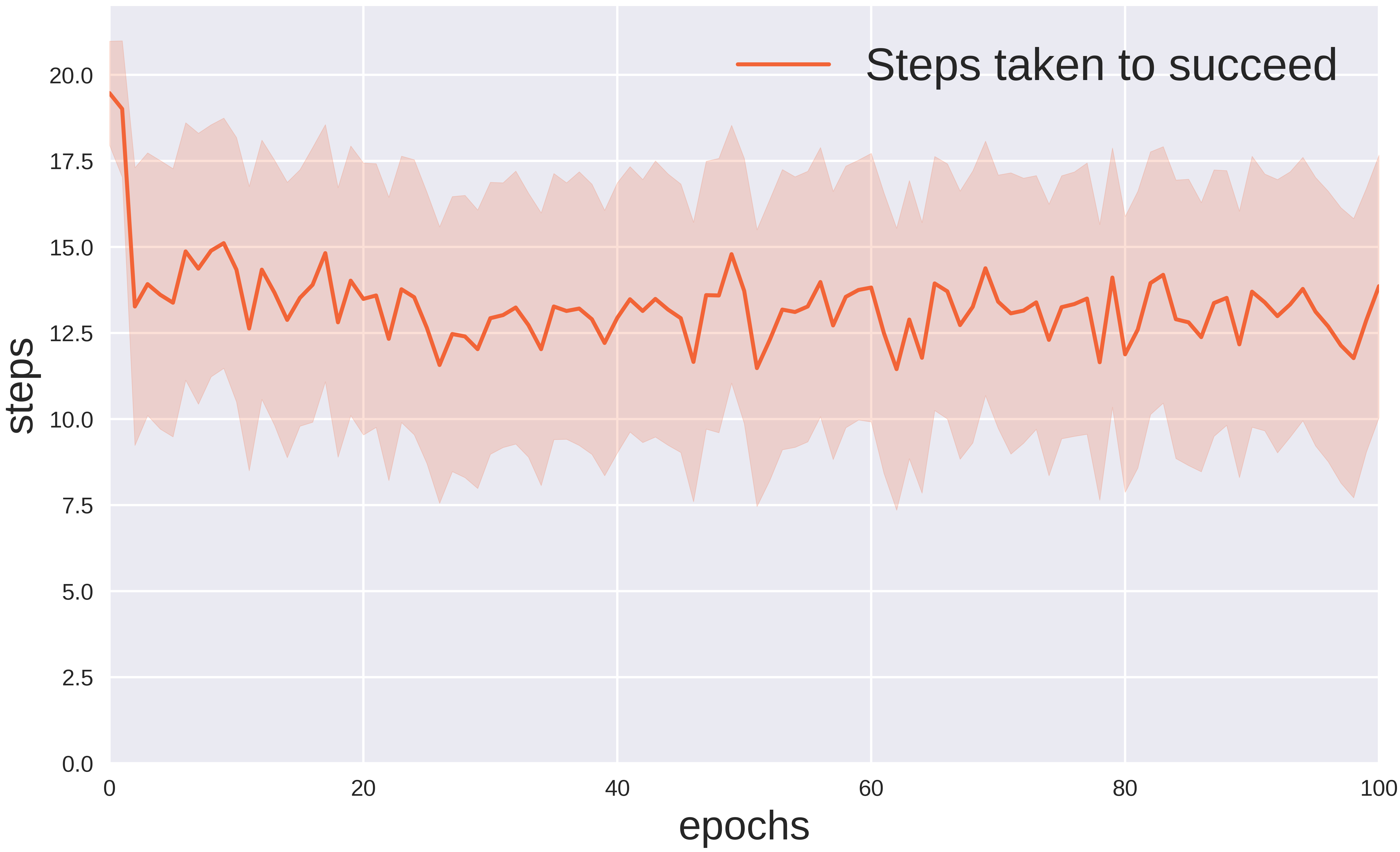}
\caption{\label{fig:succeed}The average Success duration measured during training epochs.}
\end{figure}

During the training phase, The Success rate and the Ball loss duration are expected to increase gradually as the robot learns to adjust its viewpoint through more observations and keep the ball in the viewpoint.
Figure~\ref{fig:successrate} and \ref{fig:miss} show that the robot has learned to achieve the best viewpoint and keep the ball in its point of view.
However, the Success duration is expected to decrease.
According to our reward function, the robot receives a negative reward if it doesn't move its head through the best viewpoint and in order to maximise the cumulative reward, it should learn to reach the best viewpoint as fast as possible.
Figure~\ref{fig:succeed} shows that the robot also has learned to achieve the best viewpoint as fast as possible.
We have considered each 300 time steps of the training phase as a training epoch and measured these criteria in each epoch.

\subsection{Comparison with entropy-based method}
\label{sec:Comparison}
In this section, the effect of localisation error on the performance of the proposed method is assessed against the entropy-based method. The entropy-based methods have had the best performance in active vision tasks so far in RoboCup and similar contexts as far as we know.

In this experiment we report the average Success rate in different episodes starting from different random positions as the self-localisation error increases.
The error shows the amount of inaccuracy in both position(in meter) and direction(in radian) of the robot.

Considering that localisation error is practically inevitable in a soccer playing field, just performing the action that entropy-based methods output will not be desirable for our viewpoint optimisation task.

\begin{figure}[t!]
\centering
\includegraphics[width=0.45\textwidth]{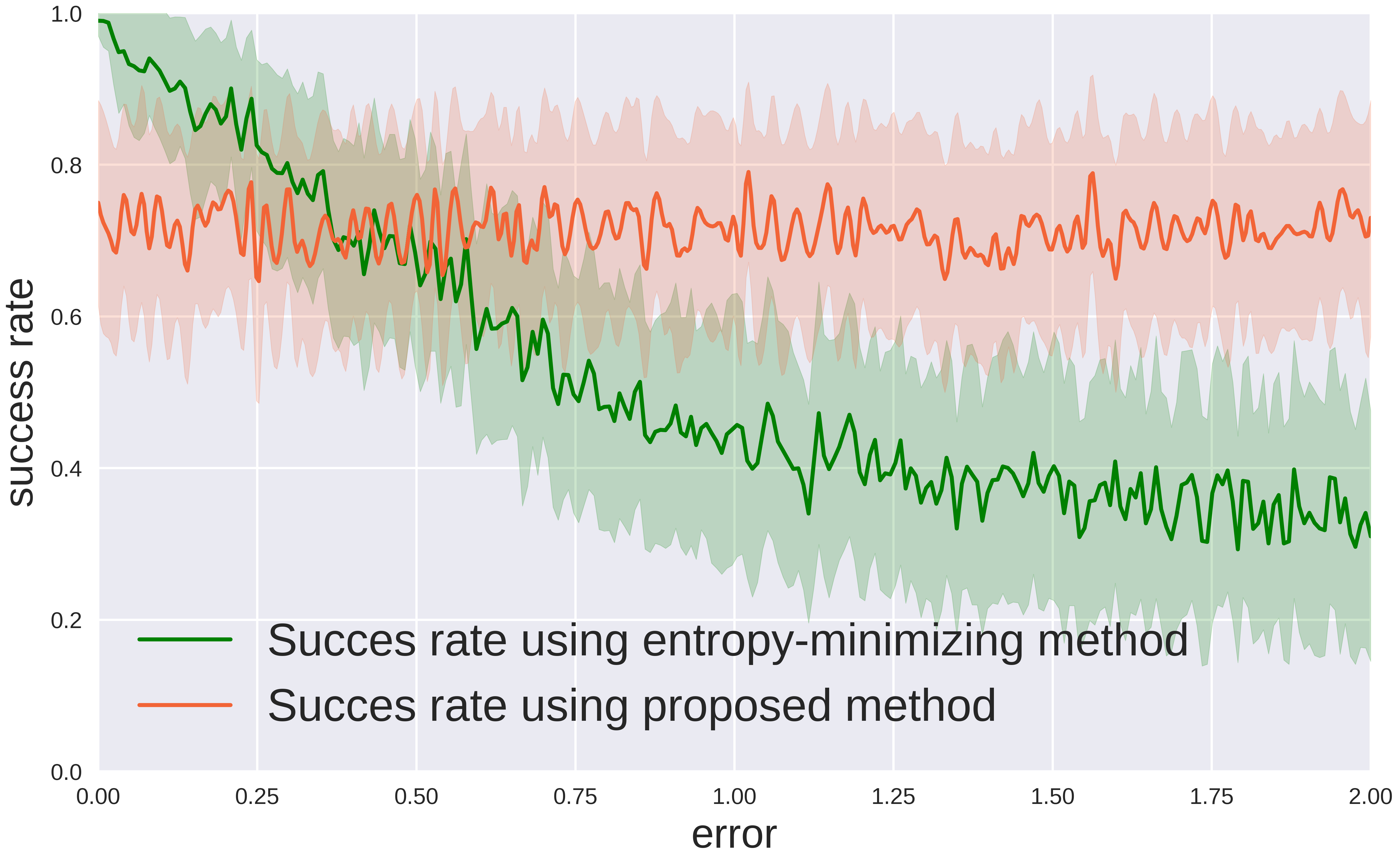}
\caption{\label{fig:Error}Comparison of our method's average Success rate against the entropy based method as the self-localisation error increases.}
\end{figure}

Figure \ref{fig:Error} illustrates the results of the experiment.
As shown in Figure \ref{fig:Error}, in entropy-based methods, the more localisation error, the lower performance can be expected.
And this is because these methods operate as a function of the robot and ball positions.
Therefore, they are highly dependent on the accuracy of the self localisation.

On the other hand, Figure \ref{fig:Error} confirms that no matter how inaccurate the localisation model is, the performance of our method remains steady.
It handles the detrimental impact of localisation error.
And this is because our proposed method works as a function of the current input image and doesn't rely on the localisation accuracy.

\vspace{-2mm}
\section{\uppercase{Conclusion}}
\label{sec:conclusion}
\vspace{-2mm}
In this work, we presented an active vision problem in the context of RoboCup which intends to control the head movements of a humanoid soccer robot.
The method formulated the problem as a Markov Decision Process using Deep Reinforcement Learning.
In the action selection phase (at the beginning of each episode), we used an entropy-minimising method applied to the UKF model which is responsible for the robot localisation.
In this work we applied the DQN algorithm.
The results of the trained model were presented and analysed.
The proposed method operates without reliance on the current belief of the environment and is compared with the previous works which only use entropy minimisation for the real-time head control.

We defined the problem as a Markov Decision Process.
Therefore the problem can be solved with newer algorithms of reinforcement learning that consider the continuous action space such as PPO \cite{schulman2017proximal} and DDPG \cite{lillicrap2015continuous}.
Also, the performance of the method might be improved by passing a rough representation of the robot position along with the image.
This can handle the problem of similarity between the symmetric observations in the soccer field.
\vspace{-2mm}

\bibliographystyle{apalike}
{\small
\bibliography{example}}

\end{document}